\documentclass[11pt,a4paper]{article}
\usepackage[hyperref]{emnlp2020}
\usepackage{times}
\usepackage{latexsym}

\usepackage{microtype}

\usepackage{graphicx}
\usepackage{subfigure}
\usepackage{amsmath}
\usepackage{CJKutf8}

\aclfinalcopy 
\def\aclpaperid{91} 

\title{{Y}ouling: an {AI}-Assisted Lyrics Creation System}
\author{Rongsheng Zhang$^{1}$\thanks{\quad Equal contribution}, Xiaoxi Mao$^{1}$\thanks{\quad Corresponding Author} \footnotemark[1], Le Li$^1$, Lin Jiang$^1$, \\
\textbf{Lin Chen$^1$, Zhiwei Hu$^1$, Yadong Xi$^1$, Changjie Fan$^1$, Minlie Huang$^2$} \\
  $^1$ {\normalsize Fuxi AI Lab, NetEase Inc., Hangzhou, China} \\
  $^2$ {\normalsize Department of Computer Science and Technology, Institute for Artifical Intelligence, State Key} \\
  {\normalsize Lab of Intelligent Technology and Systems, Beijing National Research Center for} \\
  {\normalsize Information Science and Technology, Tsinghua University, Beijing, China.} \\
  \texttt{\normalsize \{zhangrongsheng, maoxiaoxi, lile, jianglin02\}@corp.netease.com}, \\
  \texttt{\normalsize aihuang@tsinghua.edu.cn}}

\date{}

\begin{document}
\maketitle
\begin{abstract}
Recently, a variety of neural models have been proposed for lyrics generation. However, most previous work completes the generation process in a single pass with little human intervention. We believe that lyrics creation is a creative process with human intelligence centered. AI should play a role as an assistant in the lyrics creation process, where human interactions are crucial for high-quality creation. This paper demonstrates \textit{Youling}, an AI-assisted lyrics creation system, designed to collaborate with music creators. In the lyrics generation process, \textit{Youling} supports traditional one pass full-text generation mode as well as an interactive generation mode, which allows users to select the satisfactory sentences from generated candidates conditioned on preceding context. The system also provides a revision module which enables users to revise undesired sentences or words of lyrics repeatedly. Besides, \textit{Youling} allows users to use multifaceted attributes to control the content and format of generated lyrics. The demo video of the system is available at \href{https://youtu.be/DFeNpHk0pm4}{https://youtu.be/DFeNpHk0pm4}.

\end{abstract}

\begin{CJK}{UTF8}{gbsn}
\section{Introduction}

Lyrics Generation has been a prevalent task in Natural Language Generation (NLG), due to the easy availability of training data and the value of the application. However, despite the popularity of lyrics generation, there still lacks a comprehensive lyrics creation assistant system for music creators. Previous researches \cite{castro2018combining, saeed2019creative, lu2019syllable, manjavacas2019generation, watanabe2018melody, potash2018evaluating, fan2019hierarchical, li2020rigid} and systems \cite{potash2015ghostwriter, lee2019icomposer, shen2019controlling}, are mostly model-oriented, utilizing neural networks including GAN, RNN-based or Transformer-based \cite{vaswani2017attention} sequence to sequence (Seq2Seq) models for sentence-wise lyrics generation. They complete the lyrics generation process in a single pass with specific keywords or content controlling attributes as input, involving little human intervention. However, we believe the lyrics creation process should be human intelligence centered, and AI systems shall serve as assistants, providing inspiration and embellishing the wording of lyrics. 

Therefore, we demonstrate \textit{Youling}, an AI-assisted lyrics creation system, which is designed to collaborate with music creators, help them efficiently create and polish draft lyrics. To fulfill the goal, \textit{Youling} supports interactive lyrics generation, in addition to the traditional one pass full-text generation. Interactive lyrics generation allows users to carefully choose desirable sentences from generated candidates conditioned on preceding context line by line. Preceding context can be either pre-generated, written by users, or a mix. \textit{Youling} also has a revision module, which supports users to revise any unsatisfied sentences or words of draft lyrics repeatedly. 

To ensure the controllability of generated lyrics, \textit{Youling} supports multifaceted controlling attributes to guide the model to generate lyrics. These controlling attributes can be divided into two categories, content controlling attributes and format controlling attributes. Content controlling attributes include the lyrics' text style, the emotion or sentiment expressed in the lyrics, the theme described in the lyrics, and the keywords expected to appear in the lyrics. Format controlling attributes include the acrostic characters(letters), the rhymes of the lyrics, the number of sentences, and the number of words per sentence. 

\begin{figure*}[htbp]
    \centering
    \includegraphics[width=450px]{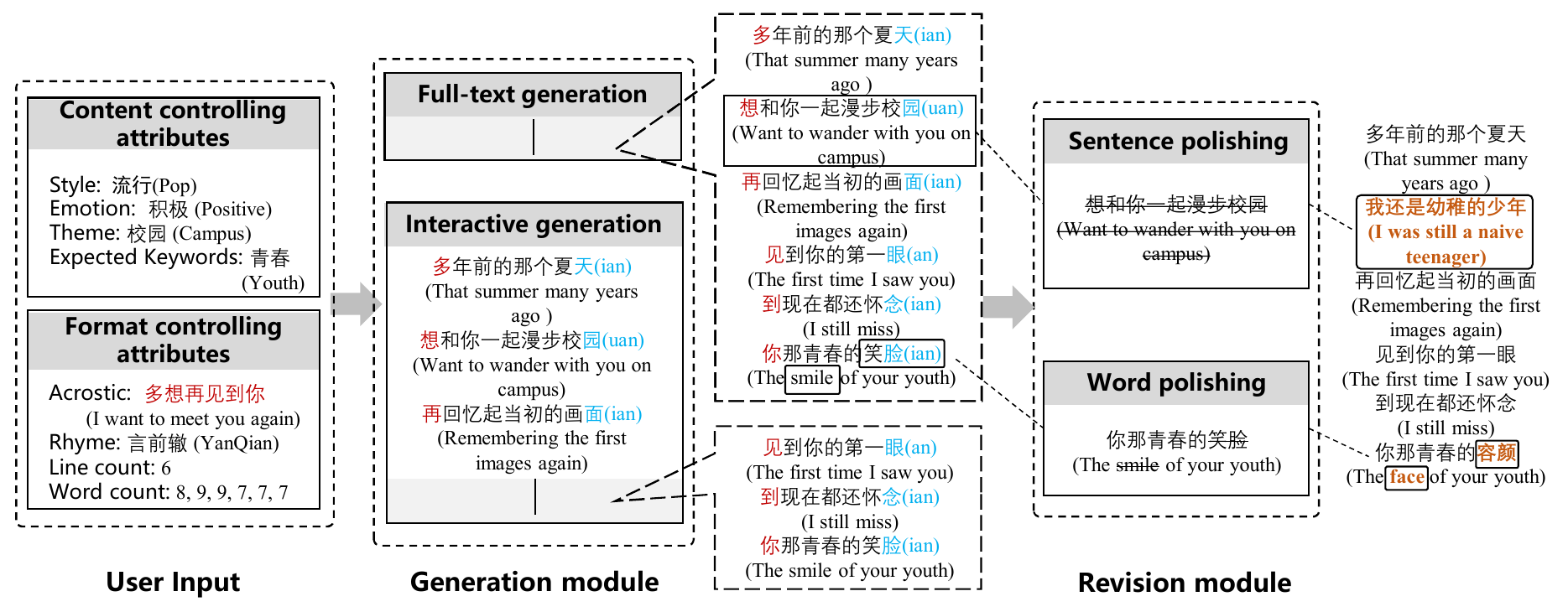}
    \caption{Architecture of \textit{Youling}. The system supports multifaceted controlling attributes in user input to control the content and format of lyrics. The generation module provides two modes for draft lyrics creation: full-text generation and interactive generation. The former generates a full lyrics while the latter generates following sentences conditioned on the preceding context. Besides, a revision module is introduced to polish undesirable sentences or words.}
    \label{fig:architecture}
\end{figure*}

To ensure the quality and relevance of generated lyrics with controlling attributes, we implement \textit{Youling} basing on a GPT-2 \cite{radford2018improving, radford2019language} based language model with 210M parameters, pre-trained on around 30 gigabytes of Chinese books corpus. We further finetune \textit{Youling} on a corpus of 300K lyrics collected online. 

The contributions of the \textit{Youling} system are summarized as follows:
\begin{enumerate}
  \item \textit{Youling} provides multiple modes to assist users in lyrics creation. It supports both the traditional one pass full-text generation and the interactive lyrics generation. It also provides a revision module for users to revise undesirable sentences or words of draft lyrics repeatedly. 
  \item To the best of our knowledge, \textit{Youling} supports the largest variety of content controlling attributes and format controlling attributes to date.
  \item \textit{Youling} is implemented on top of GPT-2 model to ensure the quality and relevance of generated lyrics with controlling attributes.
\end{enumerate}

We believe that \textit{Youling} \footnote{Our system is available at \href{https://yl.fuxi.netease.com/}{https://yl.fuxi.netease.com/}, visitors can log in with the public account (\textit{youlingtest@163.com}) and password (\textit{youling666}).} can assist music creators in lyrics creation and inspire other developers to make practical solutions for real-world problems. The 2-minute demonstration video can be available at \href{https://youtu.be/DFeNpHk0pm4}{https://youtu.be/DFeNpHk0pm4}.

\section{Architecture}

The framework of \textit{Youling} is shown in Figure \ref{fig:architecture}. The system mainly contains three parts: \textbf{user input}, \textbf{generation module} and \textbf{revision module}. We will describe them in detail in the following subsections.

\subsection{User Input}
The input includes a wide variety of controlling attributes provided by users. They can be divided into two categories: content controlling attributes and format controlling attributes. Content controlling attributes consist of the lyrics' text style, the emotion expressed in the lyrics, the theme described in the lyrics, and the keywords expected to appear in the lyrics. Our system supports four kinds of text styles, including Pop, Hip-hop, Chinese Neo-traditional and Folk; three kinds of emotion (positive, negative, and neutral); 14 kinds of themes such as college life, unrequited love, reminiscence, friendship, and so on. Format controlling attributes consist of the acrostic characters (letters), the rhymes of the lyrics, the number of lines of lyrics, and the number of words per line. Users can choose rhyme from 13 Chinese traditional rhyming groups (十三辙). 

\subsection{Generation Module}
Once users have prepared the controlling attributes, the generation module can generate lyrics in full-text generation mode or interactive generation mode. Below we will explain in detail how we implement the lyrics generation conditioned on so many controlling attributes.

\subsubsection{Full-Text Generation}
\textbf{Model and Pre-training:}
We use a Transformer-based sequence to sequence model for the generation of lyrics. To ensure the performance, we use a pre-trained language model based on GPT-2 to initialize the weights of the Transformer encoder and decoder. Our encoder uses a unidirectional self-attention similar to GPT-2; in addition, GPT-2 has only one self-attention block per layer, so the two self-attention blocks in each decoder layer share the same weights. For saving memory, the encoder and decoder share the same weights \cite{zheng2019pre}. Our pre-trained language model has 16 layers, 1,024 hidden dimensions, 16 self-attention heads, and 210 million parameters. It is pre-trained on around 30 gigabytes of Chinese Internet novels collected online, which is tokenized with Chinese character. The vocabulary size is 11,400 and the context size is 512.

\textbf{Training:}
Here we describe how we train the sequence to sequence model. We collected 300K lyrics from the Internet as training data, including 60M tokens in total. To achieve controllable generation, we need to annotate the style and mood tags corresponding to each song's lyrics and extract the keywords in the lyrics. The style tags corresponding to the lyrics were also obtained as we crawled the lyrics, so no additional processing is required. To get emotion labels, we used a ternary emotion classifier to classify emotion for each song's lyrics. The emotion classifier was trained on 20k labeled data and achieved 80\% accuracy on the validation set. To get the keywords contained in each song's lyrics, we extracted all the nouns, verbs, and adjectives in the lyrics.

After the pre-processing described above, we have the style tags, emotion tags, and keyword lists corresponding to each song's lyrics and can start building the training data. 
The encoder input is a concatenation of the style tag, emotion tag and keywords corresponding to the song lyrics with the \textit{[SEP]} special character. 
Since there are too many keywords extracted from a song's lyrics, we augment training examples by sampling different numbers of keywords multiple times. This approach is to allow the model to better generalize to the number of keywords.
To construct the decoder output, we use a special token \textit{[SEP]} to concatenate every line in a song lyrics, where the last character of each line is placed at the beginning for rhyming control. Finally, we append a special token \textit{[EOS]} to the end of the decoder output.
Kindly note that the constraints on format attributes, as well as the theme tag, are imposed during inference, so they will not be included in the training phase. 

\begin{figure}[t]
    \centering
    \includegraphics[width=220px]{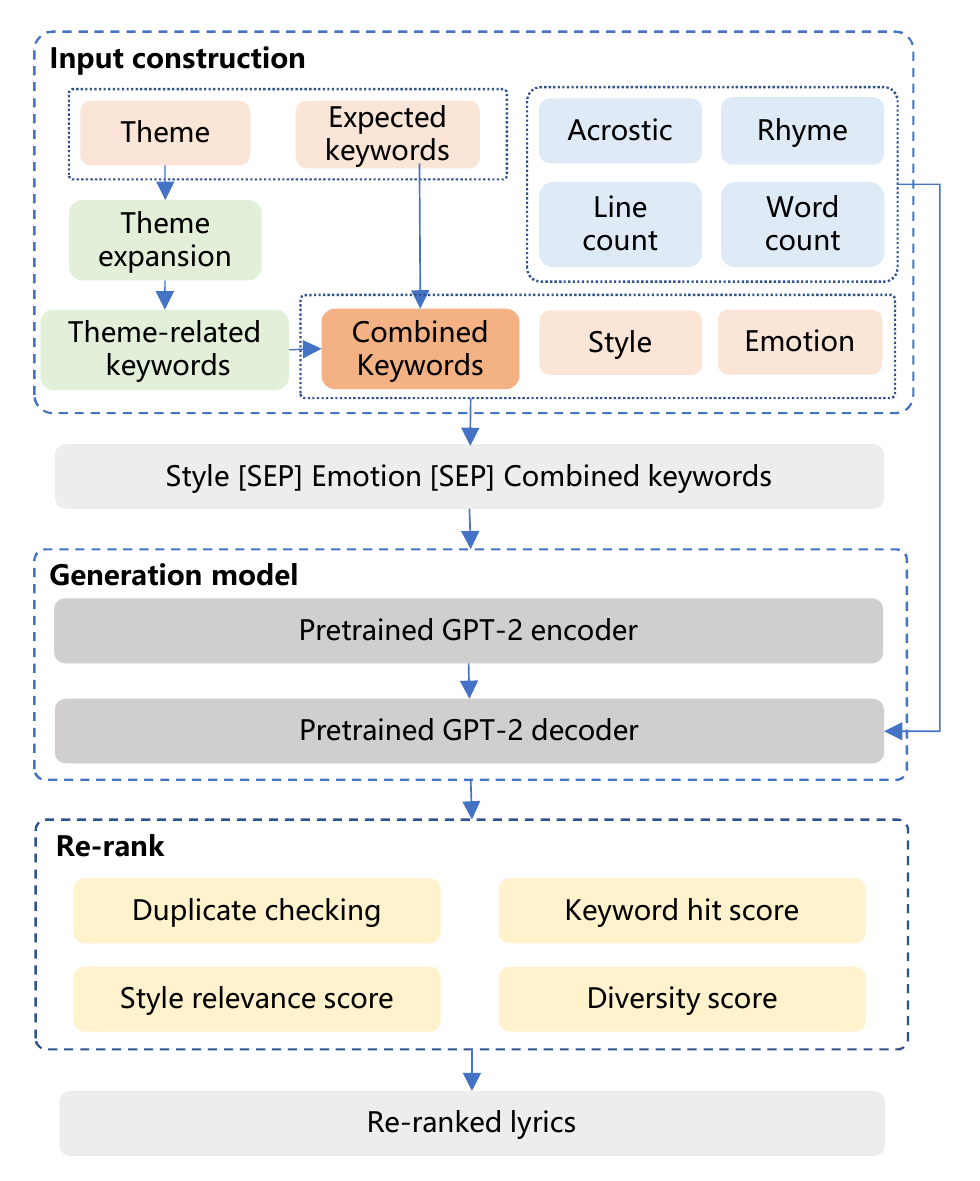}
    \caption{The inference process of full-text generation.}
    \label{fig:composing}
\end{figure}
\textbf{Inference:}
Here we introduce the inference process, as shown in Figure \ref{fig:composing}. Under full-text generation mode, the source sequence is a concatenation of the user-entered style tag, emotion tag, and keywords. The keywords include the expected keywords, as well as keywords related to the theme selected by the user. The keywords related to different themes are obtained through offline computation. We calculated PMI (Pointwise Mutual Information) for all word pairs in the lyrics corpus after removing low-frequency words. The PMI of word pair $w_i$, $w_j$ is calculated as
\begin{equation}
    \text{PMI}(w_i, w_j) = \text{log}\frac{p(w_i, w_j)}{p(w_i)*p(w_j)},
    \label{eq:omi}
\end{equation}
where $p(w_i)$ and $p(w_i, w_j)$ are the word frequency and co-occurrence frequency. We keep all word pairs with PMI above a specific threshold, which gives us the lists of keywords corresponding to specific themes. At inference time, we randomly sample the keywords list corresponding to the theme selected by the user to get the input keywords, which are then concatenated with user-entered keywords, style tag, and emotion tag to form the final source sequence. 

\textbf{Format control in decoding:}
We describe the details of format control in decoding. 
To keep the number of lines and words per line in accordance to the user's requirements, we record the number of lines and words of the generated lyrics at every decoding step and adjust the logits of \textit{[SEP]} and \textit{[EOS]} in the decoder output accordingly.
To achieve rhyming control, we always generate the last character of a line first and then generate the rest from left to right. We adjust the training examples accordingly, as mentioned before. To achieve the acrostic control, we simply amplify the corresponding logit in the decoder output to a very large value when generating the acrostic character of each line of lyrics.

\textbf{Re-rank:} We adopt the top-k sampling method at decoding to generate candidate results. Then we re-rank the candidates according to four rules. 
(1) Duplicate checking: Due to the strong copy ability of Transformer \cite{lioutas2019copy}, the generated lyrics may contain original pieces of text in the training corpus, which will introduce copyright issues. To avoid that, we remove any candidate result containing three or more lines overlapping with the training corpus. 
(2) \underline{K}eyword \underline{h}it (kh) score: For each candidate, we compute the keyword hit score as $S_{kh} = n/n_{max}$, where $n$ is the number of keywords appearing in the current candidate, $n_{max}$ is the number of of keywords in the one with the most hits in all candidates.
(3) \underline{S}tyle \underline{r}elevance (sr) score: This score measures how well each candidate matches its target style $style_{t}$. To compute the score, we train a style classifier ${g}$ on the collected lyrics corpus, and take the classification probability of the target style of the generated lyrics as $S_{sm} = g(style_{t}|lyric)$.
(4) \underline{Div}ersity (div) score: As mentioned before, the Transformer model is likely to copy original lyrics in the training data. Besides, repetition is also common in lyrics; thus, the learned model may constantly generate repeated pieces of text. Sometimes repetition is good, but too much repetition needs to be avoided. We count the number of repeated sentences in each candidate and calculate the diversity score as $S_{div} = 1 – n_{rep}/n_{tot}$, where $n_{rep}$ and $n_{tot}$ denotes the number of repeated sentences and all sentences respectively.
The final ranking score of each candidate is computed as 
\begin{equation}
    S_{rank} = \lambda_1 S_{kh} + \lambda_2 S_{sm} + \lambda_3 S_{div}, 
\end{equation}
where $\lambda_1$, $\lambda_2$, $\lambda_3$ are weights and default to $1.0$.

\subsubsection{Interactive Generation}
For the interactive generation, we use the same model used for full-text generation. The differences exist at decoding. The first difference is that under the interactive generation mode, generation is conditioned on both the encoder input and the preceding context. In other words, the interactive generation can be formulated as 
\begin{equation}
    s_{i+1}, ..., s_{i+k} = \text{Model}(X, s_0,s_1, ..., s_i), 
\end{equation}
where the $s_i$ means the \textit{i}-th line of the lyrics text $Y$, and $k$ is the number of lines to be generated. In comparison the full-text generation is just formulated as $Y = \text{Model}(X)$. The second difference is that the interactive generation mode generates only a few lines $s_{i+1}, .., s_{i+k}$ rather than the full lyrics $Y$. Hence, under the interactive generation mode, the preceding context must be provided, which can either be pre-generated by \textit{Youling}, written by the user, or a mix of them. 

For the example of interactive generation in Figure \ref{fig:architecture}, the system generates the following three lines ``\textit{The first time I saw you [SEP] I still miss [SEP] The smile of your youth}" based on the user input and the preceding context ``\textit{That summer many years ago  [SEP] Want to wander with you on campus [SEP] Remembering the first images again}".

\subsection{Revision Module}
\begin{figure}[t]
    \centering
    \setlength{\belowcaptionskip}{-0.3cm}
    \includegraphics[width=220px]{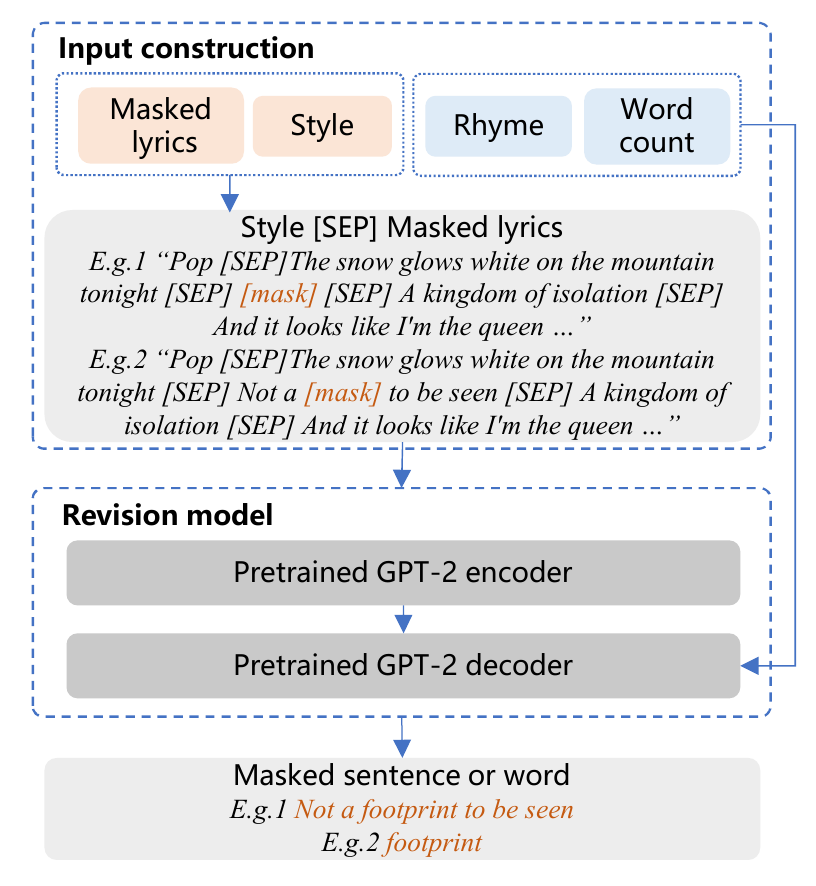}
    \caption{The process of the revision module polishing lyrics.}
    \label{fig:revising}
\end{figure}

The revision module provides useful features allowing users to further polish draft lyrics at the sentence or word level. The framework of the revision module is shown in Figure \ref{fig:revising}.

The model of revision module follows the same sequence to sequence framework used in the full-text generation model, initialized with weights of the same pre-trained language model. 

To build training examples for the model, we simply randomly replace a sentence or a word of lyrics in the training corpus with a special token \textit{[MASK]}. The result is concatenated with the corresponding style tag as the final source sequence, with the form "\textit{Style [SEP] Masked Lyrics}." The sentence or word replaced becomes the target sequence. We use an example to illustrate this idea, given the lyrics ``\textit{The snow glows white on the mountain tonight [SEP] Not a footprint to be seen [SEP] A kingdom of isolation [SEP] And it looks like I'm the queen ...}", we replace the sentence ``\textit{Not a footprint to be seen}" or the word ``\textit{footprint}'' with the masking token \textit{[MASK]}, and take the masked contents as the target sequence, as shown in Figure \ref{fig:revising}. Note that we don't treat word-level and sentence-level replacement differently, so the revision is executed with the same model.

\section{Demonstration}

\begin{figure}[t]
    \centering
    \includegraphics[width=180px]{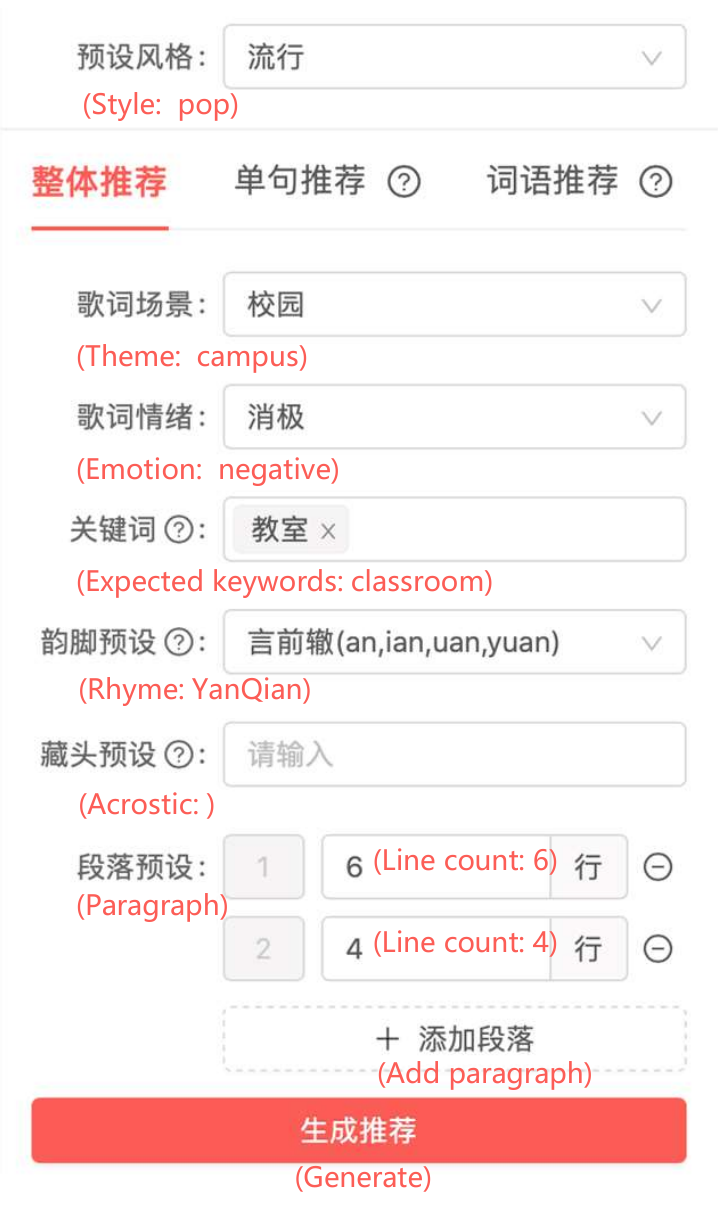}
    \caption{A case of the input page. Users can set content and format controlling attributes.}
    \label{fig:input_page}
\end{figure}
In this section, we demonstrate how \textit{Youling} assists music creators to create lyrics conveniently. 

\begin{figure*}[htbp]
\centering
\subfigure[Full-text generation]{
\begin{minipage}[t]{0.6\linewidth}
\centering
\includegraphics[width=270px]{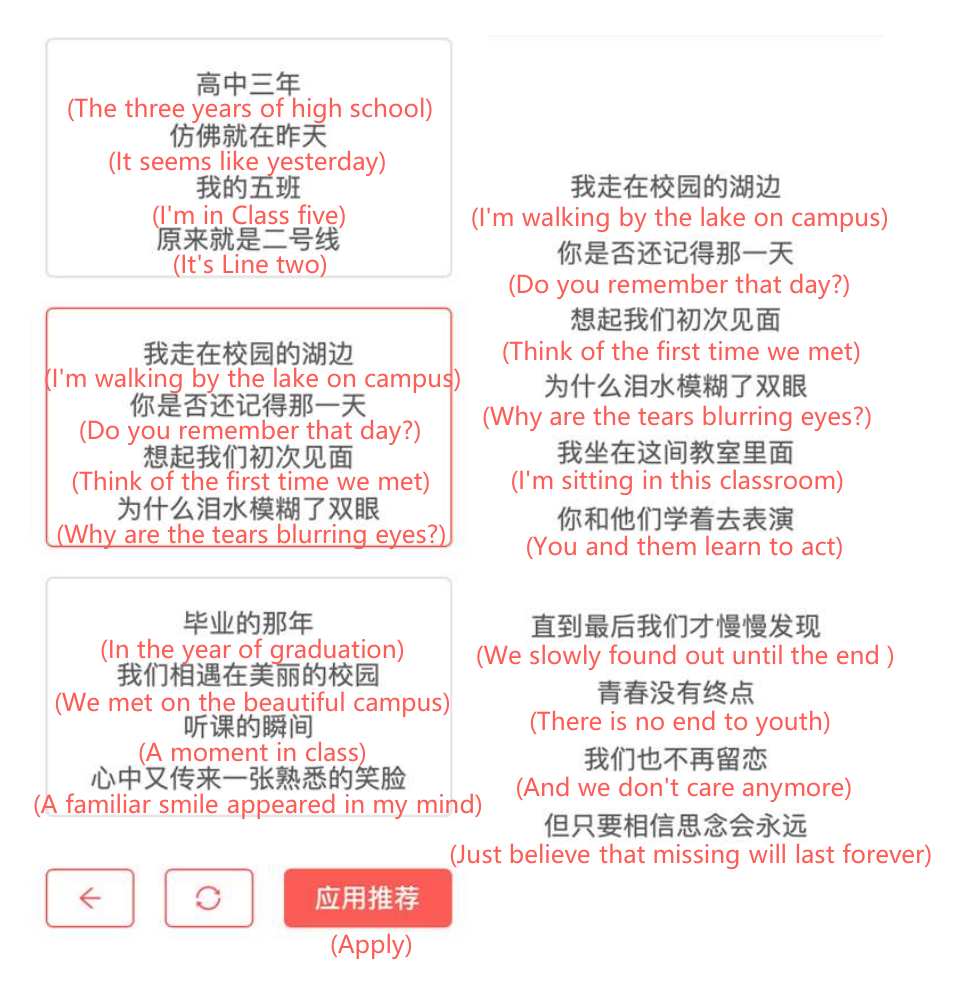}
\end{minipage}%
}%
\subfigure[Interactive generation]{
\begin{minipage}[t]{0.4\linewidth}
\centering
\includegraphics[width=180px]{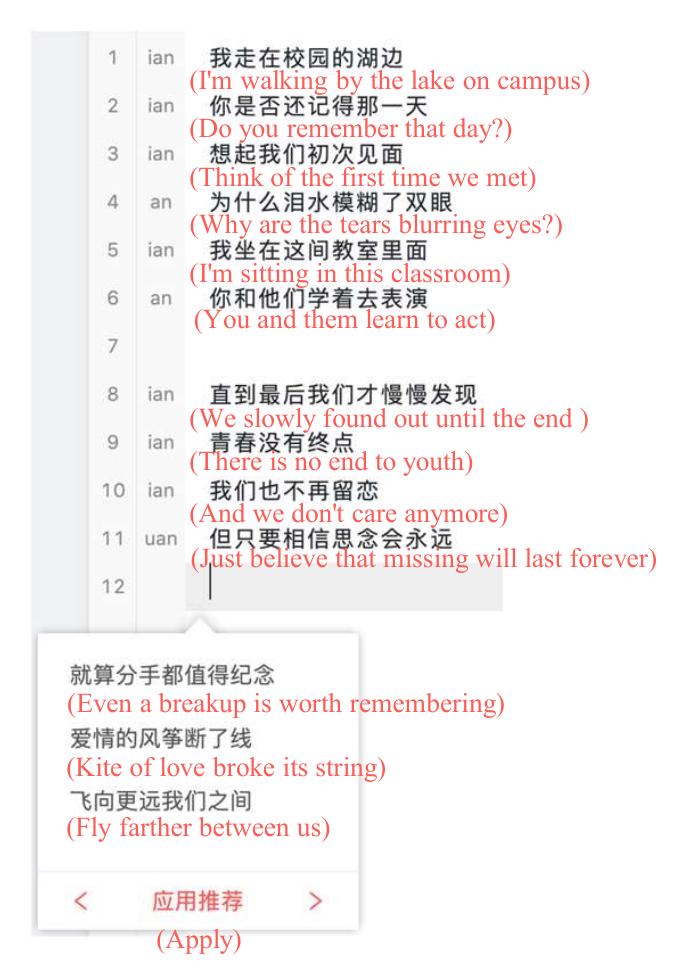}
\end{minipage}%
}%
\centering
\caption{Examples of the two generation modes. (a) Full-text generation: this mode will generates three full-text candidates for users to choose. (b) Interactive generation: the mode generates following three sentences conditioned on the preceding context. }
\label{fig:composing_page}
\end{figure*}

First, we show how to generate draft lyrics based on multifaceted controlling attributes. Users are asked to specify the controlling attributes, as shown in Figure \ref{fig:input_page}. After the controlling attributes have been prepared, we use the full-text generation mode to generate the draft lyrics, as shown in Figure \ref{fig:composing_page}(a).

After the draft lyrics are generated, we use the interactive generation mode to generate the following lines. Note that in real cases, users can directly write lyrics or modify pre-generated lyrics in the input box and generate the following lines with interactive generation mode. Here we use the unchanged generated draft lyrics for convenience of demonstration. 

\begin{figure*}[htbp]
\centering
\subfigure[Sentence polishing]{
\begin{minipage}[t]{1.0\linewidth}
\centering
\includegraphics[width=450px]{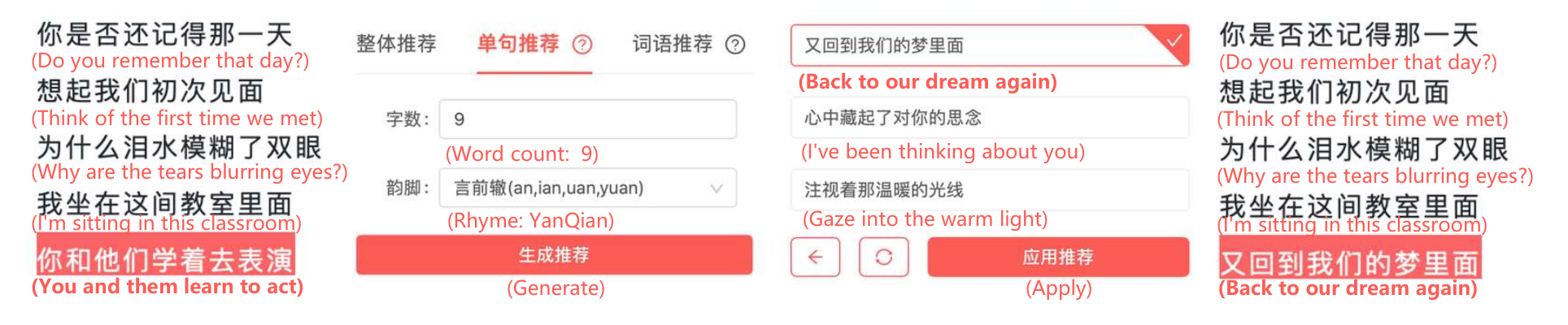}
\end{minipage}%
}%

\subfigure[Word polishing]{
\begin{minipage}[t]{1.0\linewidth}
\centering
\includegraphics[width=400px]{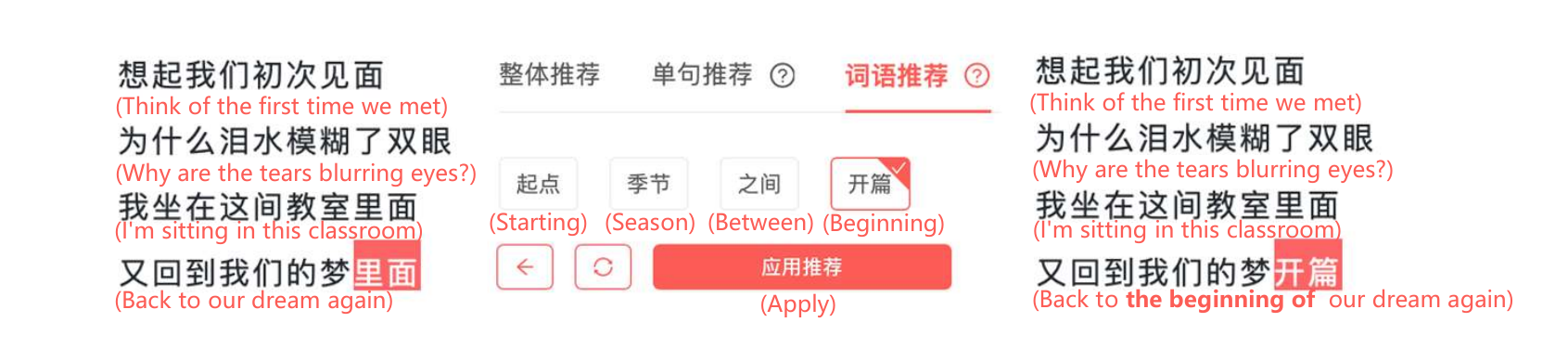}
\end{minipage}%
}%
\centering
\caption{Examples of revision module polishing sentences and words in lyrics. Users can select undesirable sentences or words, then ask the system to generate candidates for selected contents conditioned on the context.}
\label{fig:revising_page}
\end{figure*}

After completing the draft lyrics by carefully choosing the final line from generated candidates, we can further polish the undesired parts of the generated lyrics. Here we replace a flawed sentence with the best suggestion made by the revision module under sentence level, as seen in Figure \ref{fig:revising_page}(a). However, we are still not completely satisfied with the last word in the previous suggested sentence. We switch to word level and replace the last word with an appropriate word suggested by the revision model, as shown in Figure \ref{fig:revising_page}(b).

As described above, users can repeatedly revise the lyrics until desirable results are obtained. To facilitate the process, \textit{Youling} provides version control so that users can create lyrics with peace of mind.

\section{Conclusion}
In this paper, we demonstrate \textit{Youling}, an  AI-assisted lyrics creation system. \textit{Youling} can accept multifaceted controlling attributes to control the content and format of generated lyrics. In the lyrics generation process, \textit{Youling} supports traditional one pass full-text generation mode as well as an interactive generation mode. Besides, the system also provides a revision module which enables users to revise the undesirable sentences or words of lyrics repeatedly. We hope our system 
can assist music creators in lyrics creation and inspire other developers to make better solutions for NLG applications.

\bibliography{anthology,emnlp2020}
\bibliographystyle{acl_natbib}
\end{CJK}
\end{document}